\documentclass[letterpaper]{article} 
\usepackage{aaai23}  
\usepackage{times}  
\usepackage{helvet}  
\usepackage{courier}  
\usepackage[hyphens]{url}  
\usepackage{graphicx} 
\usepackage{natbib}  
\usepackage{caption} 
\usepackage{algorithm}
\usepackage{algorithmic}
\usepackage{amsmath}
\usepackage{amsthm}
\usepackage{mathtools}
\usepackage{amsfonts}       
\usepackage{scalerel,stackengine} 
\stackMath

\newcommand\reallywidehat[1]{%
\savestack{\tmpbox}{\stretchto{%
  \scaleto{%
    \scalerel*[\widthof{\ensuremath{#1}}]{\kern-.6pt\bigwedge\kern-.6pt}%
    {\rule[-\textheight/2]{1ex}{\textheight}}
  }{\textheight}%
}{0.5ex}}%
\stackon[1pt]{#1}{\tmpbox}%
}

\usepackage{todonotes}

\newcommand{\ourmethod}[1]{\textsc{ballad}}
\newcommand{\R}{\mathbb{R}}
\newcommand{\N}{\mathbb{N}}
\newcommand{\Pp}{\mathbb{P}}
\newcommand{\E}{\mathbb{E}}
\usepackage{booktabs}
\newcommand{\rr}[1]{\text{\textregistered}}
\usepackage{newfloat}
\usepackage{listings}
\DeclareCaptionStyle{ruled}{labelfont=normalfont,labelsep=colon,strut=off} 
\floatstyle{ruled}
\newfloat{listing}{tb}{lst}{}
\floatname{listing}{Listing}
\usepackage{hyperref}
\title{How to Allocate your Label Budget? Choosing between \\ Active Learning and Learning to Reject in Anomaly Detection}

\author {
    Lorenzo Perini,\textsuperscript{\rm 1}
    Daniele Giannuzzi, 
    Jesse Davis \textsuperscript{\rm 1}
}
\affiliations {
    \textsuperscript{\rm 1} KU Leuven, Department of Computer Science, DTAI \& Leuven.AI, B-3000 Leuven, Belgium\\
    lorenzo.perini@kuleuven.be, danielegiannuzzi1998@gmail.com, jesse.davis@kuleuven.be
}

\begin{document}

\maketitle

\begin{abstract}
Anomaly detection attempts at finding examples that deviate from the expected behaviour. Usually, anomaly detection is tackled from an unsupervised perspective because anomalous labels are rare and difficult to acquire. 
However, the lack of labels makes the anomaly detector have high uncertainty in some regions, which usually results in poor predictive performance or low user trust in the predictions. One can reduce such uncertainty by collecting specific labels using Active Learning (AL), which targets examples close to the detector's decision boundary.
Alternatively, one can increase the user trust by allowing the detector to abstain from making highly uncertain predictions, which is called Learning to Reject (LR). One way to do this is by thresholding the detector's uncertainty based on where its performance is low, which requires labels to be evaluated.
Although both AL and LR need labels, they work with different types of labels: AL seeks strategic labels, which are evidently biased, while LR requires i.i.d. labels to evaluate the detector's performance and set the rejection threshold. Because one usually has a unique label budget, deciding how to optimally allocate it is challenging.
In this paper, we propose a mixed strategy that, given a budget of labels, decides in multiple rounds whether to use the budget to collect AL labels or LR labels. The strategy is based on a reward function that measures the expected gain when allocating the budget to either side. We evaluate our strategy on 18 benchmark datasets and compare it to some baselines.
\end{abstract}

\section{Introduction}~\label{sec:introduction}
Anomaly detection is the task of automatically detecting examples that do not follow expected patterns~\cite{chandola2009anomaly}. These examples, named anomalies, are usually indicative of critical events such as water leaks in stores~\cite{perini2022transferring}, breakdowns in gas turbines~\cite{zhao2016review}, or failures in the petroleum extraction~\cite{marti2015anomaly}. Such critical events usually come along with elevated (maintenance) costs or with substantial natural damages (e.g., dispersion of petroleum or gas). Thus, detecting anomalies in time is a relevant task that limits such resource waste. 

Collecting labels, especially for anomalies, is often a hard task because anomalies are costly events (e.g., machine failures cannot be voluntarily induced), or simply time-consuming (e.g., you may need to label 100s of examples before getting an anomaly). Thus, anomaly detection is often tackled from an unsupervised perspective. However, the lack of labels usually forces the unsupervised detector to have high uncertainty on specific regions of the example space~\cite{perini2020quantifying}. High uncertainty is undesirable because it is often associated with poor predictive performance or reduced trust in the predictions.

This uncertainty can be tackled in two complementary ways. On the one hand, one can try to learn a more accurate detector by acquiring a limited number of labels using Active Learning (AL)~\cite{abe2006outlier}. 
On the other hand, it is possible to increase the user trust in the detector's outputs by allowing the detector to abstain from making a prediction when it is highly uncertain, which is called Learning to Reject (LR)~\cite{hendrickx2021machine,de2000reject}. One way to do this is to set a rejection threshold on the detector's uncertainty based on where its performance is poor~\cite{cortes2016learning}. However, evaluating the detector performance requires labels.

Both of these approaches rely on labeled data. However, the types of labels needed for each approach are quite different.  
Many AL strategies rely on biased sampling strategies such as explicitly targeting acquiring labels, for example, for which the detector is highly uncertain (i.e., near the detector's current decision boundary) as these are known to yield better performance~\cite{pimentel2020deep,culotta2005reducing}.
Alas, using such labels to evaluate the detector's performance, as required when setting the threshold in LR, will yield a biased performance estimate and hence a sub-optimal threshold~\cite{marrocco2007empirical}. Thus, if a user has a fixed budget for acquiring labels there is a tension between collecting (a) strategic labels that can be used to train a better detector, or (b) i.i.d. labels that can be used to evaluate performance and set a proper rejection threshold. Therefore, a data scientist is confronted with the challenging question of how they should optimally allocate their label budget between these two purposes.

In this paper, we assume that the label budget can be split and allocated in multiple rounds. We introduce \ourmethod{} (\underline{B}udget allocation for \underline{A}ctive \underline{L}earning and \underline{L}earning to reject in \underline{A}nomaly \underline{D}etection) a novel adaptive strategy that, in each allocation round, (1) measures the potential reward obtained by assigning the budget to either AL or LR, and (2) chooses the highest reward option to collect the labels.

\section{Preliminaries and Related Work}~\label{sec:related_work}

\paragraph{\textbf{Anomaly Detection.}} Let $X$ be a $d-$dimensional random variable with unknown $p(X)$. We are given a dataset $D = \{x_1, \dots, x_n\}$ with $n$ examples and $d$ features is drawn i.i.d. from $p(X)$. Let $V = \{x_{n+1}, \dots, x_m\} \sim_{i.i.d} p(X)$, $m > n$, be a validation set. Let $Y$ be the label random variable, such that $Y|X=x$ indicates the class label ($1$ if anomaly, $0$ if normal) for $x \in \R^d$. An anomaly detection problem is the task of finding an anomaly score function $h\colon \R^d \to \R$ and a threshold $t\in \R$ such that $Y = h^{(t)}(X)$, where $h^{(t)}(x) = 1$ if $h(x)\ge t$, $0$ otherwise. Usually, one sets $t$ based on the contamination factor $\gamma$, i.e. the proportion of anomalies~\cite{perini2022estimating}.

\paragraph{\textbf{Pool-based Active Learning (AL).}} 
The goal of pool-based AL strategies is to reduce the detector's uncertainty by selecting the most informative training instances.
The AL approaches can be classified into $3$ categories~\cite{monarch2021human}: \emph{uncertainty-based} sampling strategies aim to select the unlabeled data samples with the highest uncertainty~\cite{hacohen2022active}, \emph{diversity strategies} capture the diversity among the training data~\cite{abe2006outlier,dagan1995committee}, \emph{combined strategies} integrate the advantages of uncertainty-based and diversity-based
criteria~\cite{ebert2012ralf}. 

\paragraph{\textbf{Learning to Reject (LR).}} The goal of a detector's reject option is to abstain from making a prediction when a detector is too uncertain about predicting a test example~\cite{hendrickx2021machine,cortes2016learning}. Our goal is to develop a detector-agnostic strategy that does \emph{ambiguity rejection}, as novelty rejection would reject all anomalies. Thus, we use a \emph{dependent} rejector architecture~\cite{chow1970optimum}. We indicate by $\mathcal{C}(x)$ the detector's confidence for predicting $x \in V$, and with $\tau \in [0,1]$ the rejection threshold. If the confidence is below $\tau$, the prediction is rejected $h^t(x) = \rr{}$. \emph{Note that for appropriate inference, we need to collect validation labels randomly (i.i.d.)}.


\section{A strategy to allocate the label budget}\label{sec:method}
\noindent This paper tackles the following problem:
\begin{description}
\item[Given:] initially unlabeled training set $D$ and validation set $V$, the dataset's contamination factor $\gamma$, an anomaly detector $h$, and a label budget $B$;
\item[Do:] decide whether, in each allocation round $k$, to acquire labels for $D$ (AL) or for $V$ (LR).
\end{description}
Both training the detector with more labels (AL) and learning a threshold using larger validation data (LR) improve the detector's performance. However, choosing the side to maximize such improvement is challenging for multiple reasons.
First, it requires measuring the reward of either side, i.e. the expected gain in terms of the detector's improvement.
Second, the rewards need to be on a similar scale such that neither side is privileged during the process.
Third, comparing a standard detector to one with the reject option is challenging because the latter needs ad-hoc metrics to overcome the problem of predicting three classes (anomaly, normal, reject)~\cite{nadeem2009accuracy}.

In this paper, we introduce \ourmethod{}, a strategy that measures the reward of allocating the budget for AL, i.e. collecting \emph{strategic} labels on the training set, and for LR, i.e., collecting \emph{random} labels on the validation set. 
Let $B = k \cdot b \in \N$ be our labelling budget. We perform $k$ rounds and the labels of $b$ examples are queried in each round. 
We \underline{initialize} the problem by (1) training the detector with no labels and setting a default rejection threshold, and (2) collecting $b$ random labels for $V$ (LR) and for $D$ (AL) for a total of $2b$ labels. This allows us to compute the \underline{initial rewards} by measuring how the detector varies from (1) to (2): for LR, we measure the variation after re-setting the validation threshold; for AL, we measure the variation after re-training the detector with the new labels. Then, \underline{we start the allocation loop}. In each round, we allocate the budget to the option (LR or AL) with the highest reward, and we update the reward using the new labels. We propose two alternative reward functions: 1) the \emph{entropy} reward looks at the detector's probabilities, either for prediction (AL), or for rejection (LR); 2) the \emph{cosine} reward considers the predicted class labels, either anomaly yes/no (AL), or rejection yes/no (LR).



\subsection{Measuring the reward}
Because we do not know how beneficial the next label allocation would be for the detector, we look at the past and measure the effect of the last allocation round. Our challenge is to design a reward function that reflects the gain when querying the labels. We use the following methods to derive the reward for both AL and LR, by using as detector's probabilities either the probability of predicting anomaly (AL), or the probability of rejecting the example (LR). Similarly to~\citet{vercruyssen2022multi}, we consider two scenarios:
\paragraph{Entropy.} Adding more labels has the ability to decrease the overall uncertainty of the anomaly detector. Thus, we measure the variation of the detector's probabilities as:
\begin{equation}~\label{eq:reward_entropy}
    \mathcal{R}^e(k) = \E_{x\sim X}\left[\left| \mathcal{H}(h_k(x)) - \mathcal{H}(h_{k-1}(x))\right|\right],
\end{equation}
where $\mathcal{H}(h(x)) = - p \log_2 p$ is the entropy of the detector's probabilities $p$, and the subscript indicates the query round (for $k > 2$). A large difference in entropy means a large detector variation, which indicates a large impact of the new labels and, in turn, a large reward $\mathcal{R}^e$.
\paragraph{Cosine.} More directly, one can measure the impact of the labels in terms of variation of class predictions. Given the detector's probabilities, we threshold them at $0.5$ and assign value $1$ to higher probabilities and $0$ to lower ones. Thus, we measure the cosine similarity between different outputs as
\begin{equation}\label{eq:reward_cosine}
    \mathcal{R}^c(k) = \E_{D\sim X}\left[ 1 - \frac{h_k(D) \cdot h_{k-1}(D)}{\|h_k(D)\|\cdot \|h_{k-1}(D)\|}
    \right],
\end{equation}
where $h(D)$ is a vector containing the outputs ($0$ or $1$) by the detector $h$, and $\|\cdot\|$ is the Euclidean norm. This metric is less sensitive to little variations in the detector and discriminates more in case the new labels change the predicted class.

\subsection{Deriving the detector's probabilities}
Measuring the reward needs some probabilities, which are not easy to derive due to the partially supervised setting. For both prediction and rejection, we exploit the squashing function: given a positive real score $s\in \R_+$ and a threshold $\lambda \in \R_+$, the squashing function 
$$\mathcal{S}_{\lambda}\colon \R_+ \to (0,1),\quad \mathcal{S}_{\lambda}(s) = 1- 2^{-\frac{s^2}{\lambda^2}}$$ maps $s$ to a probability $>0.5$ if $s > \lambda$, and $\le 0.5$ otherwise.
Roughly speaking, $\mathcal{S}_{\lambda}$ calibrates the probabilities by centering $\lambda$ as the decision threshold.

\paragraph{Detector's posterior probabilities.} Given the contamination factor $\gamma$, a common approach to set the threshold $t$ is by forcing the detector to have a training positive class rate equal to $\gamma$. Thus, one can center the probabilities to $t$ by transforming the anomaly scores $h(x)$ through the squashing function:
\begin{equation*}
    \Pp(h^t(x) = 1) = \mathcal{S}_{t}(h(x))\ \text{ s. t. } \ t = Q_{h}(1-\gamma).
\end{equation*}
We set $t$ as the $1-\gamma$th quantile of the score distribution such that only a proportion of $\gamma$ scores have $\Pp(h^t(x) = 1) \ge 0.5$.

\paragraph{Rejection probabilities.} Given a validation set with some labels, we (1) set a specific detector confidence $\mathcal{C}(x)$, and (2) set the rejection threshold $\tau \in [0,1]$. For the former, we use the detector's posterior probabilities:
\begin{equation*}
    \mathcal{C}(x) = 2\times \left|\Pp(h^t(x) = 1) - 0.5 \right| \in [0,1].
\end{equation*}
Thus, the closer $\Pp(h^t(x) = 1)$ is to $0.5$ (high uncertainty), the lower the detector confidence. For the latter, we optimize the threshold $\tau$ over the validation set (only the labeled examples) by minimizing a cost function $\mathcal{M}(h^t)$. Finally, we compute the rejection probabilities by centering 1-confidence values to the rejection threshold, i.e. by applying the squashing function
\begin{equation*}
    \Pp(h^t(x) = \rr{}) = \mathcal{S}_{\tau}(1-\mathcal{C}(x)).
\end{equation*}

\subsection{The cost-based evaluation metric}
Given a detector with a reject option and a detector without it,
we cannot compare their performance on the non-rejected examples, as they would have different test sets. Thus, we introduce a cost-based evaluation metric. Formally, given a rejection cost $c_r>0$, a false positive cost $c_{fp}>0$, and a false negative cost $c_{fn} > 0$, the detector is evaluated as:
\begin{equation*}
\begin{split}
    \mathcal{M}_h &= c_r \cdot \Pp(h^t(X) = \rr{}) + c_{fp} \cdot \Pp(h^t(X) = 1 | Y = 0) \\
    &+ c_{fn} \cdot \Pp(h^t(X) = 0 | Y = 1).
\end{split}
\end{equation*}
Note that we assume cost null for the correct predictions, while every misprediction as well as the rejection gets penalized. Because rejecting is assumed to be less costly than mispredicting, the rejection cost needs to satisfy the inequality $c_r \le min \{c_{fp} \times (1-\gamma), c_{fn} \times \gamma \}$, otherwise one could predict either always normal and pay an expected cost of $c_{fn} \times \gamma$, or always anomaly and pay $c_{fp} \times (1-\gamma)$.

\section{Experiments}\label{sec:experiments}
We experimentally answer the following questions:
\begin{itemize}
    \item[Q1.] Does \ourmethod{} result in lower costs when compared to using only AL or LR?
    \item[Q2.] Which reward metric is better?
    \item[Q3.] Is the reward function on a similar scale for AL and LR?
    \item[Q4.] How does our strategy behave when varying $c_{fp}$, $c_{fn}$ ?
\end{itemize}

\subsection{Experimental setup}
\paragraph{Methods.} We compare \ourmethod{}\footnote{Code available at https://github.com/Lorenzo-Perini/Ballad} to two baselines: \textsc{All-in-AL} allocates all the budget for active learning and sets the rejection threshold using the (biased) training labels; on the contrary, \textsc{All-in-LR} allocates all the budget for learning to reject and uses an unlabeled training set.

\paragraph{Data.}

\begin{table}[htbp]
\caption{Properties of the $18$ datasets used.}
\label{tab:data}
\centering
\begin{tabular}{lrrr}
\toprule
Dataset & \# Examples & \# Features & $\gamma$\\
\midrule
ALOI             &12384      &27          &0.0304 \\
Annthyroid       &7129       &21          &0.0749 \\
Arrhythmia       &271        &259         &0.0996 \\
Cardiotocography &1734       &21          &0.0496 \\
Glass            &214        &7           &0.0421 \\
InternetAds      &1682       &1555        &0.0499 \\
KDDCup99         &48113      &40          &0.0042 \\
PageBlocks       &5473       &10          &0.1023 \\
PenDigits        &9868       &16          &0.0020 \\
Pima             &526        &8           &0.0494 \\
Shuttle          &1013       &9           &0.0128 \\
SpamBase         &2661       &57          &0.0499 \\
Stamps           &340        &9           &0.0912 \\
WBC              &223        &9           &0.0448 \\
WDBC             &367        &30          &0.0272 \\
WPBC             &160        &33          &0.0562 \\
Waveform         &3443       &21          &0.0290 \\
Wilt             &4655       &5           &0.0199 \\
\bottomrule
\end{tabular}
\end{table}

We carry out our study on $18$ publicly available benchmark datasets, which are widely used in the literature~\cite{campos2016evaluation}. See Table~\ref{tab:data} for the properties.

\paragraph{Setup.}
For each of the 18 benchmark datasets, we go as follows:
(i) we split the dataset into training, validation and test sets using the proportions $40-40-20$ (we have a large validation set to better measure the impact of rejection);
(ii) we fit the anomaly detector on the unlabeled dataset and set the rejection threshold to the default value of $0.1$;
(ii) we allocate a budget $b$ to LR and AL by randomly selecting the initial examples;
(iii) we optimize the rejection threshold and measure the LR reward;
(iv) we train the anomaly detector on the partially labeled training set and measure the AL reward;
(v) we allocate the next round budget $b$ to the option with the highest reward and repeat (iii) or (iv) until the whole budget $B$ is used. During each of the steps, we measure the detector performance on the test set using our cost function. We set $B$ to the $30\%$ of the training set's size, and $b$ to $2\%$ of it, such that we run $15$ allocation rounds. We repeat (i - v) $10$ times and report the average results. In total we run $18 \times 15 \times 10 = 2700$ experiments.

\paragraph{Costs and hyperparameters.} We set $c_{fp} = c_{fn} = 1$ and $c_r = \gamma$, following the cost inequality. \textsc{ssdo}~\cite{vercruyssen2018semi} with its default parameters is used as the semi-supervised anomaly detector~\cite{soenen2021effect}. We use \textsc{IForest}~\cite{liu2008isolation} as its unsupervised prior. We use Uncertainty Sampling as the active learning strategy~\cite{zhan2021comparative}, and the entropy as default reward. For setting the rejection threshold, we use Bayesian Optimization (\textsc{gp$\_$minimize} implemented in \textsc{Skopt}) with $20$ calls~\cite{frazier2018bayesian} and limit the rejection rate on the validation set to $50\%$.

\subsection{Experimental results}

\begin{figure*}[htbp]
\centering
\includegraphics[width = .99\textwidth]{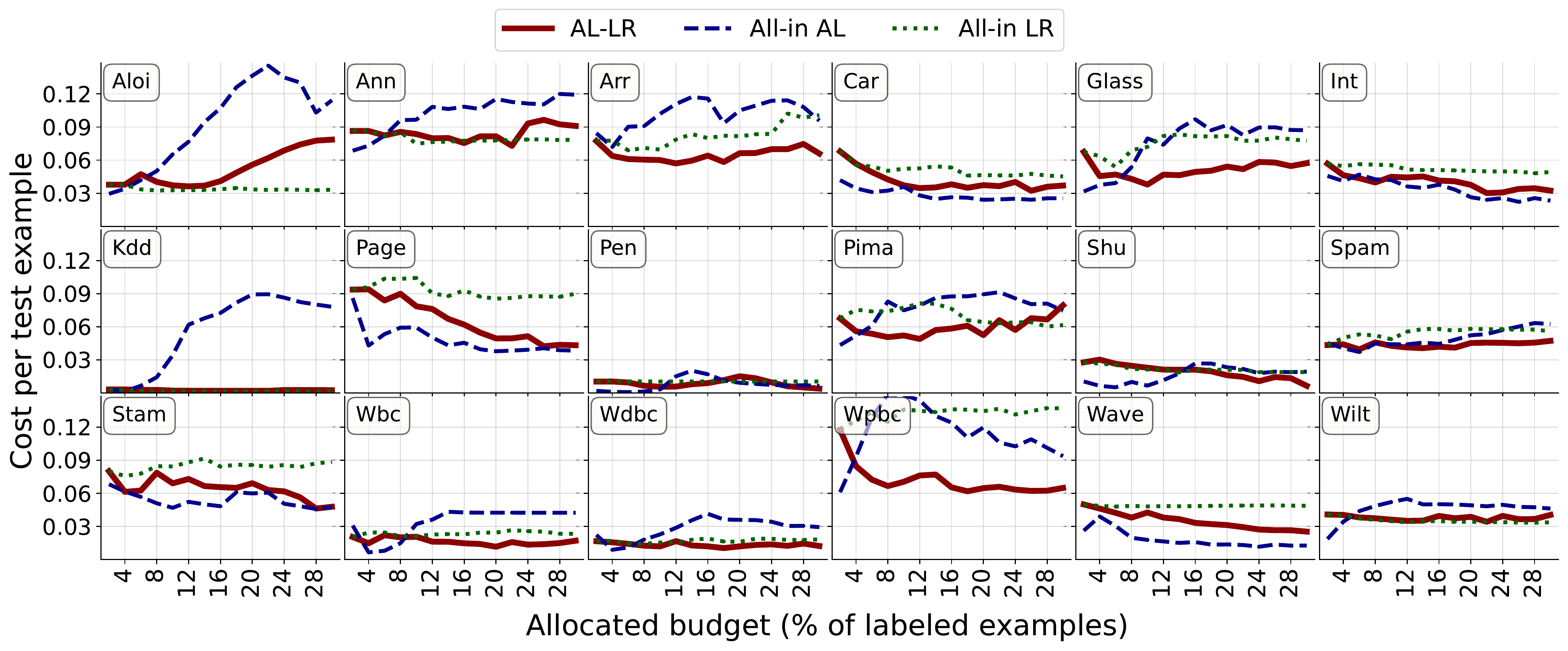}
\caption{Comparison between \ourmethod{} and the \textsc{All-in} strategies on the 18 benchmarks. The x-axis reports the $15$ rounds of $2\%$ labels each. The y-axis shows the average cost per test example. \ourmethod{} obtains lower costs in the majority of cases.
}
\label{fig:Q1_comparison}
\end{figure*}

\textbf{Q1. Comparing \ourmethod{} to the \textsc{All-in} strategies.} 
Figure~\ref{fig:Q1_comparison} shows the comparison between \ourmethod{} with the entropy reward and the \textsc{All-in} strategies on the $18$ benchmark datasets. 
On $8$ datasets (Arrhythmia, Glass, KddCup99, Pima, SpamBase, Wbc, Wdbc, Wpbc), \ourmethod{} results in evident lower costs, although sometimes the difference is small. On $5$ datasets (Cardiotocography, InternetAds, PageBlocks, Stamps, Waveform) \ourmethod{} performs similar/worse than \textsc{All-in-AL}. This happens because \textsc{ssdo} has an overall high performance and a contained uncertainty in the predictions. On the other hand, in $3$ cases (Aloi, Annthyroid, Wilt), allocating all the budget for LR has a lower cost. This is due to the detector being inaccurate and unable to learn from the training labels, which makes learning an optimal threshold more convenient.  As support for this intuition, we analyze the plain test AUC of \textsc{SSDO} on the whole test set (no rejection) for each of the three previous cases. By aggregating over the rounds, \textsc{SSDO} obtains an average AUC equal to $0.86$, $0.88$, and $0.57$ when the best strategy is, respectively, \ourmethod{}, \textsc{All-in-AL}, and \textsc{All-in-LR}. Finally, \ourmethod{} obtains an overall average cost of $0.043$, which is $\approx 20\%$ lower than the baselines' average cost ($0.055$ for \textsc{All-in-AL}, $0.054$ for \textsc{All-in-LR}).

\begin{table}[htpb]
\centering
\begin{tabular}{ccc}
\toprule
 Budget&  Entropy $\mathcal{R}^e$ &  Cosine $\mathcal{R}^c$\\
\midrule
2\%           &  0.0536 $\pm$ 0.0401 &  \bf{0.0399 $\pm$ 0.0303} \\
4\%           &  0.0465 $\pm$ 0.0330 &  \bf{0.0411 $\pm$ 0.0334} \\
6\%           &  0.0443 $\pm$ 0.0284 &  \bf{0.0398 $\pm$ 0.0321} \\
8\%           &  0.0436 $\pm$ 0.0303 &  \bf{0.0399 $\pm$ 0.0306} \\
10\%          &  0.0420 $\pm$ 0.0299 &  \bf{0.0411 $\pm$ 0.0325} \\
12\%          &  \bf{0.0416 $\pm$ 0.0303} &  0.0433 $\pm$ 0.0347 \\
14\%          &  \bf{0.0413 $\pm$ 0.0306} &  0.0448 $\pm$ 0.0367 \\
16\%          &  \bf{0.0408 $\pm$ 0.0288} &  0.0456 $\pm$ 0.0372 \\
18\%          &  \bf{0.0403 $\pm$ 0.0290} &  0.0457 $\pm$ 0.0355 \\
20\%          &  \bf{0.0412 $\pm$ 0.0297} &  0.0451 $\pm$ 0.0363 \\
22\%          &  \bf{0.0407 $\pm$ 0.0301} &  0.0451 $\pm$ 0.0359 \\
24\%          &  \bf{0.0421 $\pm$ 0.0325} &  0.0438 $\pm$ 0.0361 \\
26\%          &  \bf{0.0417 $\pm$ 0.0345} &  0.0438 $\pm$ 0.0363 \\
28\%          &  \bf{0.0416 $\pm$ 0.0332} &  0.0438 $\pm$ 0.0380 \\
30\%          &  \bf{0.0418 $\pm$ 0.0345} &  0.0427 $\pm$ 0.0354 \\
\bottomrule
\end{tabular}
\caption{Average ($\pm$ std) cost per test example over the datasets grouped by allocation round for each of the two reward functions. For low budgets, the cosine reward obtains lower costs, while not being competitive for high budgets.}
\label{tab:rewards}
\end{table}

\noindent \textbf{Q2. Which reward function works better?} We analyze both types of reward functions that we introduced in Eq.~\ref{eq:reward_entropy} and Eq.~\ref{eq:reward_cosine}. Table~\ref{tab:rewards} shows the mean and standard deviation of the cost, divided by allocation round. Overall, using the cosine reward builds a strategy that produces on average low costs for little budget ($\le 10\%$), whereas, for a higher budget, the entropy reward obtains better average costs. This is due to the highly imbalanced choices made by the cosine reward: the strategy opts for AL in $93\%$ of the cases, which usually improves a lot the detector's performance with few labels but tends to produce little effect when enough labels are given. On the other hand, the entropy reward is more balanced and opts for AL in $63\%$ of the cases. This allows the detector to keep decreasing the costs while learning during the allocation rounds and obtain more steady performance.

\noindent\textbf{Q3. Is the entropy reward balanced for AL and LR?}
\begin{figure}[htbp]
\centering
\includegraphics[width = 0.4\textwidth]{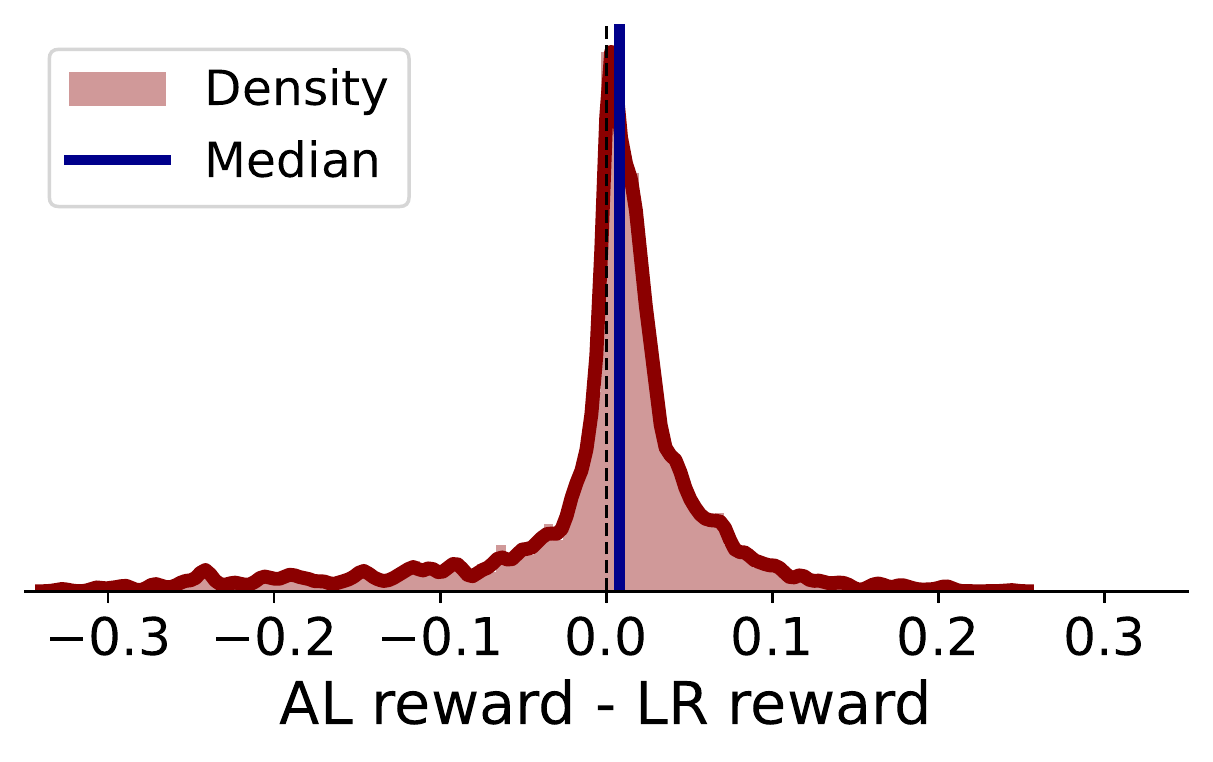}
\caption{Distribution of the difference between AL's and LR's entropy reward. The median close to $0$ indicates the absence of a predominant strategy.
}
\label{fig:reward_pdfs}
\end{figure}
Figure~\ref{fig:reward_pdfs} shows the distribution of the difference between AL and LR entropy rewards over all the $2700$ experiments. Negative values indicate that the LR reward is higher than AL's one, while the opposite holds for positive values. Overall, the median is close to $0$, which means that there is no clearly predominant strategy. Because the left tail of the density is larger than the right one, we conclude that LR rewards have higher variability (std $= 0.07$ vs $0.03$).

\noindent\textbf{Q4. The impact of varying $c_{fp}$, and $c_{fn}$.} In this experiment, we penalize more false positives and false negatives by setting, one at a time, $c_{fp}$ and $c_{fn}$ to $10$. We compare \ourmethod{} to the two \textsc{All-in} baselines. For $c_{fp} = 10$,  our strategy is still the best for low budgets ($<15\%$), reducing the relative cost by between $5\%$ and $25\%$ with respect to the runner-up \textsc{All-in-LR}. However, for higher budgets ($>15\%$), \textsc{All-in-LR} becomes the best strategy as it reduces \ourmethod{}'s cost by around $20\%$ and \textsc{All-in-AL}'s cost by more than $40\%$. This happens because the anomaly detector produces too many false positives, which, if rejected, allow us to reduce the cost. For $c_{fn} = 10$, \ourmethod{} performs much better than the baselines, reducing their cost by around $20\%$ (vs \textsc{All-in-LR}) and $24\%$ (vs \textsc{All-in-AL}).

\section{Conclusion}\label{sec:conclusion}

We proposed \ourmethod{}, a novel strategy to decide whether to allocate the budget for Active Learning (AL), i.e. labeling strategic training instances, or for Learning to Reject (LR), i.e. labeling a random validation set. Our key insight is that we can measure the expected reward when labeling either set and allocate the label in the next round to the option with the highest reward. We proposed two reward functions (entropy and cosine similarity based). Experimentally, we evaluated \ourmethod{} on $18$ datasets, and show that it performs better than simply allocating all the labels to either AL or LR.


\newpage
\paragraph{Acknowledgements.}
This work was presented at the 1st AAAI Workshop on Uncertainty Reasoning and Quantification in Decision Making (UDM23). \\
This research is supported by an FB Ph.D. fellowship by FWO-Vlaanderen (grant 1166222N) [LP], the Flemish Government under the “Onderzoeksprogramma Artificiële Intelligentie (AI) Vlaanderen” programme [JD], and KUL Research Fund iBOF/21/075 [JD].

\bibliography{bibliography}
\end{document}